\documentclass[a4paper, 10pt, conference]{ieeeconf}      % Use this line for a4 paper

\IEEEoverridecommandlockouts                              % This command is only needed if 
\usepackage{graphics} % for pdf, bitmapped graphics files
\usepackage{epsfig} % for postscript graphics files
\usepackage{mathptmx} % assumes new font selection scheme installed
\usepackage{times} % assumes new font selection scheme installed
\usepackage{amsmath} % assumes amsmath package installed
\usepackage{amssymb}  % assumes amsmath package installed
\usepackage{hyperref}
\usepackage{xcolor}
\usepackage[natbib=true,style=numeric, sorting=none,maxcitenames=40,mincitenames=4,giveninits=false, citestyle=numeric-comp]{biblatex}
\usepackage{arydshln}
\title{\LARGE \bf
M2CURL: Sample-Efficient Multimodal Reinforcement Learning via Self-Supervised Representation Learning for Robotic Manipulation
}

\author{Fotios Lygerakis$^{1}$ Vedant Dave$^{1}$ Elmar Rueckert$^{1}$% <-this % stops a space
\thanks{$^{1}$Chair of Cyber-Physical Systems, University of Leoben, Austria}%
\thanks{*Contact Email: fotios.lygerakis@unileoben.ac.at}% <-this % stops a space
}

\begin{document}

\maketitle
\thispagestyle{empty}
\pagestyle{empty}

%%%%%%%%%%%%%%%%%%%%%%%%%%%%%%%%%%%%%%%%%%%%%%%%%%%%%%%%%%%%%%%%%%%%%%%%%%%%%%%%
\begin{abstract}
One of the most critical aspects of multimodal Reinforcement Learning (RL) is the effective integration of different observation modalities. Having robust and accurate representations derived from these modalities is key to enhancing the robustness and sample efficiency of RL algorithms. However, learning representations in RL settings for visuotactile data poses significant challenges, particularly due to the high dimensionality of the data and the complexity involved in correlating visual and tactile inputs with the dynamic environment and task objectives. 
To address these challenges, we propose Multimodal Contrastive Unsupervised Reinforcement Learning (M2CURL). Our approach employs a novel multimodal self-supervised learning technique that learns efficient representations and contributes to faster convergence of RL algorithms. Our method is agnostic to the RL algorithm, thus enabling its integration with any available RL algorithm. We evaluate M2CURL on the Tactile Gym 2 simulator and we show that it significantly enhances the learning efficiency in different manipulation tasks. This is evidenced by faster convergence rates and higher cumulative rewards per episode, compared to standard RL algorithms without our representation learning approach. Project website:
{\footnotesize \url{https://sites.google.com/view/M2CURL/home}}

\end{abstract}

%%%%%%%%%%%%%%%%%%%%%%%%%%%%%%%%%%%%%%%%%%%%%%%%%%%%%%%%%%%%%%%%%%%%%%%%%%%%%%%%
\section{INTRODUCTION}
Studies in cognitive science and psychology demonstrate that human beings possess the capacity to integrate diverse informational inputs, such as visual, auditory, and tactile data, for enhanced comprehension of their physical environment and informed decision-making processes~\cite{spelke1976infants,sullivan1983infant,walker1997infants}. The multidimensional nature of human information processing, emphasizing complexity and adaptability, mirrors the objectives of Multimodal Reinforcement Learning (MRL)~\cite{mdl,srivastava2012multimodal,mml,mdf}. MRL has gained prominence, with successful applications~\cite{kahou2016emonets,zhang2018multimodal,qian2018multimodal,vstrl} underscoring its growing importance in the field.

Robust and accurate representations derived from these modalities are key to enhancing the robustness and sample efficiency of Reinforcement Learning (RL) algorithms. However, representing visuotactile data in RL settings is challenging due to the high dimensionality and complexity involved in correlating visual and tactile inputs with dynamic environments and task objectives~\cite{review1,gao2022tactile}. In tasks involving manipulation, vision is the primary sensory input for robots until they physically interact with an object or surface. Following this, tactile feedback becomes the main source of detailed information, especially in cases where the robot's arm obstructs the visual field. Thus, 
\begin{figure}[htbp]
    \centering
    \includegraphics[width=\linewidth, trim={0.25in 2.7in 0 2.7in}, clip]{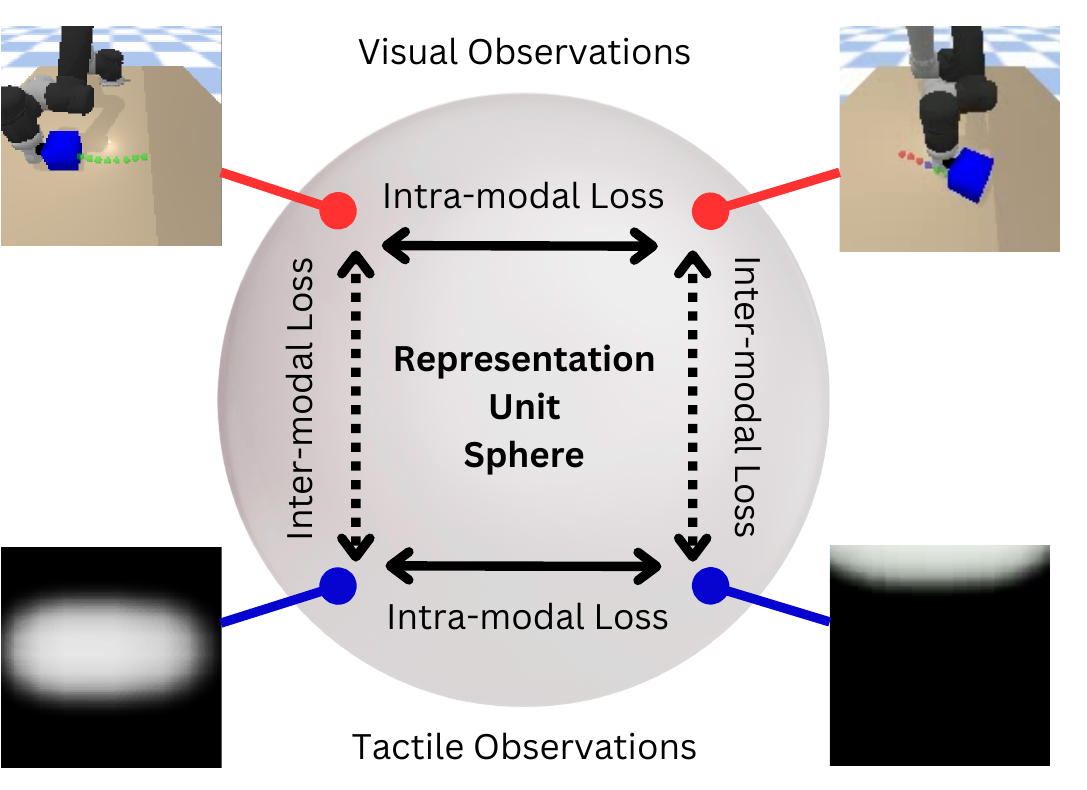}
    \caption{Representation of M2CURL features on a unit sphere of codes. This diagram illustrates the projection of visual and tactile observations from high-dimensional space onto a unit sphere, where both intra and inter-modality losses are computed. These losses form the core of the contrastive multimodal loss, essential in M2CURL's learning process.}
    \label{fig:yourlabel}
\end{figure}
sophisticated learning techniques are essential to efficiently extract and integrate key features from visual and tactile data streams, ensuring consistent performance in unpredictable and rapidly changing environments. Furthermore, the intrinsic noise and variability present in sensory data, especially tactile sensing, add complexity to the process of learning representations~\cite{ding2023adaptive,lin2023attention}. For RL algorithms, this is crucial, as the quality of representations directly influences the agent's decision-making capabilities and task success.

The integration of visual and tactile sensory information to address previously identified difficulties has become a primary focus of recent researches~\cite{liu2016visual,li2022seehearfeel,luo2018vitac,zhu2018novel,maqsood2020multi}. Advances in the realm of self-supervised learning (SSL) have been pivotal in this context, especially considering the limited availability of labeled datasets in tactile and visuotactile domains~\cite{chen2022visuotactile,kerr2023selfsupervised,yang2022touch,guzey2023dexterity}. This approach has been proved to be highly beneficial for generating meaningful representations for RL applied to manipulation tasks,  by reducing the dependency on labor-intensive labeled datasets, thus presenting a promising avenue to mitigate the hurdles in tactile data collection. Leveraging unlabeled data facilitates more efficient and robust representation learning, thereby enhancing the performance of RL algorithms in complex, real-world tasks. Despite the benefits, self-supervised learning approaches have not yet been effectively integrated with RL for executing complex manipulation tasks.

To tackle these challenges, we propose Multimodal Contrastive Unsupervised Reinforcement Learning (M2CURL). Our method facilitates the learning of intra and inter-modal representations by employing two pairs of encoders to compute four InfoNCE~\cite{CPC} losses. The within-modality (intra) losses, one for each modality, maximize the agreement among similar modality instances, while the cross-modality (inter) losses maximize the similarity between different modalities within the same sample. 
These representations are subsequently integrated into the RL algorithm, leading to accelerated convergence of the algorithm. Importantly, M2CURL is algorithm-agnostic, allowing for seamless integration into any pre-existing RL framework. We conducted experiments using the Tactile Gym 2~\cite{tactilegym2}. Our results confirm the M2CURL's effectiveness, revealing substantial improvements in learning efficiency and RL agent performance. Our findings indicate faster learning convergence and higher reward accumulation, compared to the baseline RL methods.

\section{Related Work}
The fusion of visual and tactile modalities in robotics is an expanding research domain, with numerous studies exploring its complexities. This section aims to emphasize significant contributions made in this evolving field.
 
\subsection{Unified Representation Learning for Visual and Tactile Modalities}
Recent research underscores the importance of effectively integrating visual and tactile information to create efficient representations~\cite{navarro2023visuo}. Li et al.~\cite{li2019connecting} utilized cross-modal prediction, merging visual and tactile signals via conditional adversarial networks. They improved vision-touch interaction, prevented GAN mode collapse with data rebalancing, and included touch scale and location data in their model. Additionally, Lin et al.~\cite{8793885} learned to identify objects by a cross-modality instance recognition model. Similarly, Huaping et al.~\cite{liu2016visual} devised a visual-tactile fusion framework utilizing a joint group kernel sparse coding approach to resolve the challenge of weak pairing in visual-tactile data samples. Luo et al.\cite{luo2018vitac} learned a joint latent space shared by two modalities, i.e., vision and tactile data, for the task of cloth texture recognition. In the generative domain, Zhong et al.~\cite{zhong2022touching} used Neural Radiance Fields and Conditional GANs to generate camera-based tactile observations from desired poses. Yang et al.~\cite{Yang_2023_ICCV} used the latent diffusion model to learn representations that generate images from touch and vice-versa. Recently, Dave et al.~\cite{dave2024multimodal} employed Multimodal Contrastive training to derive representations from both visual and tactile data, facilitating the performance of classification tasks across diverse datasets.

\subsection{Multimodal sensory integration for Robotic Manipulation}
Several advancements have been made in integrating multi-sensory information for improving grasping and manipulation tasks\cite{hogan2020tactile,dave2022predicting,9786532,xia2022review}. Calandra et al.~\cite{calandra2017feeling,8403291} found that incorporating tactile sensing into visual information significantly improves grasping outcomes. Lee et al.~\cite{9043710} incorporated self-supervised learning to derive representations from visual and tactile inputs. These representations were subsequently fused with optical flow and classical controllers for executing downstream manipulation tasks. Tian et al.~\cite{8794219} presented a tactile Model Predictive Control, which relies on a learned forward predictive model to execute goal-based actions. Chen et al.~\cite{chen2022visuotactile} developed the Visuo-Tactile Transformer (VTT), which combined visual and tactile data through spatial attention, showing improved efficiency in manipulation tasks. Kerr et al.~\cite{kerr2023selfsupervised} developed a self-supervised learning approach using intra-modal contrastive loss to learn representations for tasks like garment feature tracking and manipulation. However, this approach is not integrated with RL for action execution. Recently, Guzey et al.~\cite{guzey2023dexterity} demonstrated that applying self-supervised methods for learning tactile representations from a dataset of arbitrary, contact-rich interactions yielded enhanced outcomes in manipulation tasks.

\section{Background}
This section provides a brief explanation of the core principles behind our M2CURL framework, focusing on mathematical models and referencing significant algorithms in the relevant literature.

\subsection{Contrastive Learning}

Contrastive Learning is a powerful technique in self-supervised learning that focuses on learning representations by distinguishing between similar (positive) and dissimilar (negative) pairs of samples. One of the most prominent contrastive losses in use is the InfoNCE loss~\cite{CPC}. Mathematically, it can be expressed as optimizing the following loss:

\begin{equation}
L = -\log \frac{\exp(\text{sim}(z_i, z_j) / \tau)}{\sum_{k=1}^{N} \exp(\text{sim}(z_i, z_k) / \tau)}\nonumber
\end{equation}

where \( z_i, z_j \) are representations of positive pairs, \( \text{sim}(\cdot) \) denotes a similarity measure (e.g., cosine similarity), \( \tau \) is a temperature scaling parameter, and \( N \) is the number of negative samples. 

Recent advancements in contrastive learning include approaches like SimCLR \cite{simCLR} and MoCo \cite{moco}, which have set new benchmarks in unsupervised representation learning in image and language processing domains. These methods emphasize the importance of a rich set of augmentations and a large number of negative samples to learn generalizable features.

\begin{figure*}[h]
    \centering
    \includegraphics[width=\linewidth, trim={0 1in 0 0.5in}, clip]{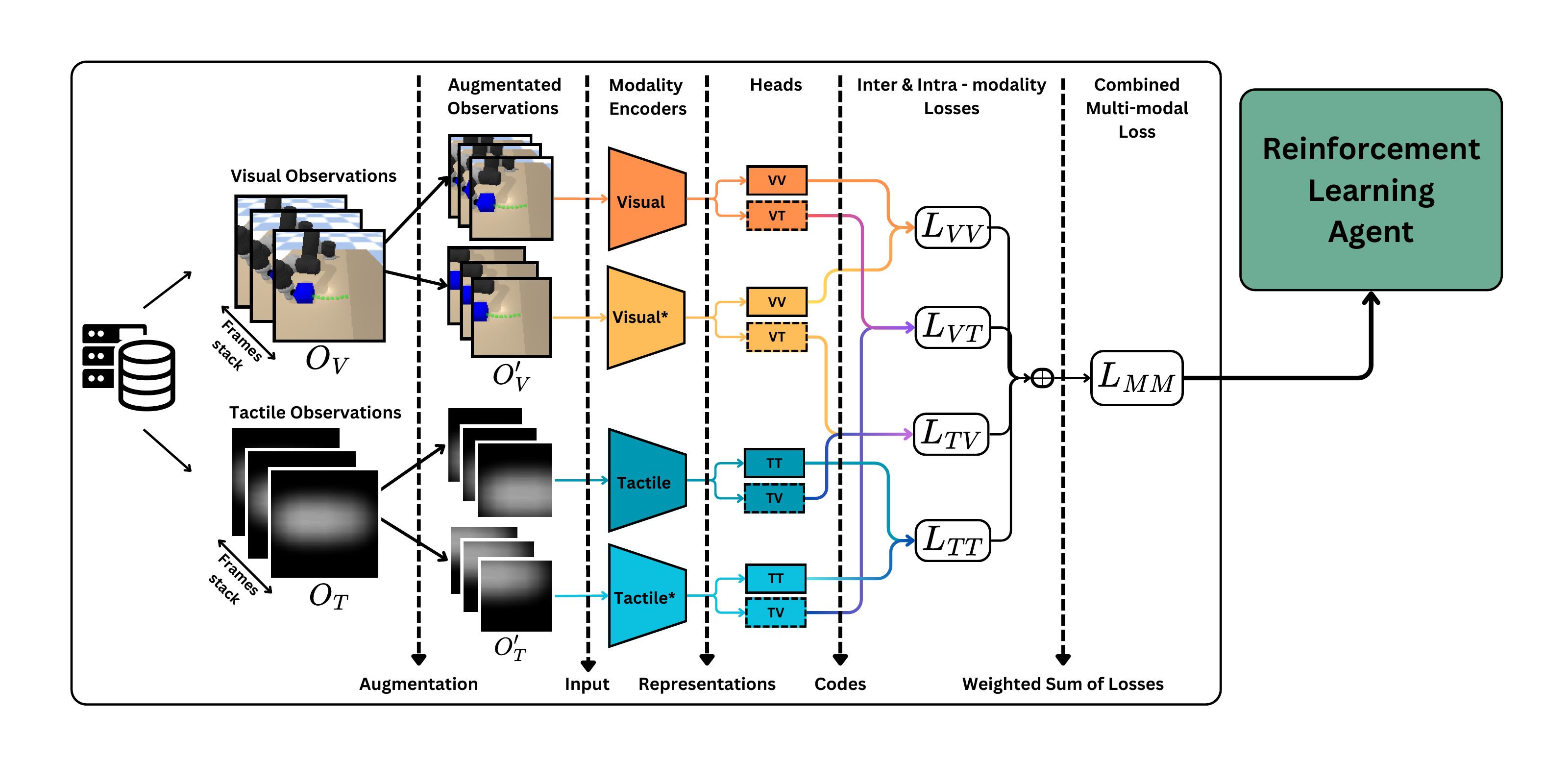}
    \caption{The M2CURL Architecture: First, a batch of visuotactile observations are sampled from the replay buffer. Then, two random augmentations are applied for the query (online) and key (momentum) encoders, and their representation is computed. The query and key representations are used to compute the inter and intra-modality codes using the respective heads, from which the different inter and intra-modality losses are computed. Finally, the weighted sum of the sub-losses is passed to the RL algorithm as a combined multimodal contrastive loss $\mathcal{L}_{MM}$. Momentum encoders are denoted with *.}
    \label{fig:architecture}
\end{figure*}
\subsection{Soft Actor-Critic Algorithm}

The Soft Actor-Critic (SAC) algorithm \cite{haarnoja2017soft} is an off-policy actor-critic framework that incorporates entropy regularization to effectively balance exploration and exploitation. Being off-policy, SAC benefits from the ability to learn from past experiences stored in a replay buffer, enhancing sample efficiency. However, this approach may present some challenges in maintaining stability compared to on-policy methods like PPO \cite{schulman2017proximal}, particularly when adapting to varying environments and during the hyperparameter tuning process. The SAC's objective function is defined as:

\begin{equation}
J(\pi) = \mathbb{E}_{(s_t, a_t) \sim \rho_\pi} \left[ \sum_{t} \gamma^t \left( R(s_t, a_t) + \alpha \mathcal{H}(\pi(\cdot | s_t)) \right) \right]\nonumber
\end{equation}

where \( \rho_\pi \) is the state-action distribution under policy \( \pi \), \( R(s_t, a_t) \) is the reward function, \( \gamma \) is the discount factor, \( \alpha \) is the temperature parameter determining the importance of the entropy term \( \mathcal{H} \), and \( \pi(\cdot | s_t) \) is the policy. SAC's strength lies in its robustness and ability to handle high-dimensional action spaces, but it can be computationally intensive due to the need for frequent policy updates.

\subsection{Proximal Policy Optimization}

Proximal Policy Optimization (PPO) \cite{schulman2017proximal} is an on-policy algorithm that optimizes a modified version of the expected return to maintain a balance between policy improvement and stability. Being on-policy, PPO updates policies using data collected by the current policy, ensuring relevance and consistency of the learning process. However, this approach can be less sample-efficient compared to off-policy methods. The PPO objective function with clipping is:

\begin{equation}
L^{CLIP}(\theta) = \mathbb{E}_t \left[ \min(r_t(\theta) \hat{A}_t, \text{clip}(r_t(\theta), 1 - \epsilon, 1 + \epsilon) \hat{A}_t) \right]\nonumber
\end{equation}

where \( r_t(\theta) = \frac{\pi_\theta(a_t | s_t)}{\pi_{\theta_{old}}(a_t | s_t)} \) is the probability ratio of the new policy to the old policy, \( \hat{A}_t \) is the advantage estimate at time \( t \), and \( \epsilon \) is a hyperparameter that controls the clipping range. PPO's main advantage is its stability and ease of implementation, but its on-policy nature requires more interactions with the environment, potentially making it slower to converge in complex tasks.

\section{Multimodal Contrastive Unsupervised Reinforcement Learning (M2CURL)}
The following sections describe the M2CURL method architecture, its novel visuotactile contrastive learning strategy, and its simple integration with off-the-shelf RL algorithms. 
\subsection{M2CURL Method Architecture}

The M2CURL architecture is tailored for processing multimodal data, with a specific focus on visual (\( \mathcal{O}_V \)) and tactile (\( \mathcal{O}_T \)) observations. It comprises two main components: the encoders and the inter/intra modality heads.

\textbf{Online Encoders:} 
There are two online encoders, \( f_{visual}(\theta_v) \) and \( f_{tactile}(\theta_t) \). The online encoders are used to extract the features used by the RL algorithm. \( f_{visual} \) is responsible for processing visual data (\( \mathcal{O}_V \)), converting it into a feature representation \( z_v \). Similarly, \( f_{tactile} \) processes tactile data (\( \mathcal{O}_T \)), resulting in a tactile feature representation \( z_t \).

\textbf{Momentum Encoders:} 
Each encoder has a corresponding momentum encoder, \( m_{visual}(\theta_{v,m}) \) and \( m_{tactile}(\theta_{t,m}) \), which are updated using a momentum-based approach from the online encoder. Their representations are solely used to compute the contrastive losses. The momentum encoding ensures stability and temporal consistency in the learned representations \cite{moco}. The momentum update rules are given by:
\begin{equation}
\theta_{v,m} \leftarrow \alpha \theta_{v,m} + (1 - \alpha) \theta_v\nonumber
\end{equation}
\begin{equation}
\theta_{t,m} \leftarrow \alpha \theta_{t,m} + (1 - \alpha) \theta_t\nonumber
\end{equation}
where \( \alpha \) is the momentum coefficient, controlling the rate of update from the primary encoders to the momentum encoders.

\textbf{Heads:} 
Each encoder has two heads, each being a 2-layer neural network. For the visual encoder, the heads are:
\begin{itemize}
        \item \textbf{Vision-to-vision head (\( H_{vv} \))}, which focuses on learning representations within the visual domain:
        \begin{equation}
        c_{vv}(z_v) = H_{vv} z_v, \quad \text{where} \quad z_v = f_{visual}(o_V)\nonumber
        \end{equation}
        \item \textbf{Vision-to-tactile head (\( H_{vt} \))}, dedicated to correlating visual features with tactile data:
        \begin{equation}
        c_{vt}(z_v) = H_{vt} z_v\nonumber
        \end{equation}

\end{itemize}
For the tactile encoder, the heads are:
\begin{itemize}
        \item \textbf{Tactile-to-tactile head (\( H_{tt} \))}, concentrating on learning within the tactile domain:
        \begin{equation}
        c_{tt}(z_t) = H_{tt} z_t, \quad \text{where} \quad z_t = f_{tactile}(o_T) \nonumber
        \end{equation}
        \item \textbf{Tactile-to-vision head (\( H_{tv} \))}, for integrating tactile features into the visual representation space:
        \begin{equation}
        c_{tv}(z_t) = H_{tv} z_t \nonumber
        \end{equation}
\end{itemize}

These components collectively enable the encoders to be trained via a contrastive learning approach, thereby developing rich and effective multimodal representations. This enhances the RL agent’s performance in environments requiring sophisticated sensory integration. 
The overall architecture, including the momentum encoders and the two-layer neural network heads, is illustrated in Figure \ref{fig:architecture}.

\subsection{Visuotactile Augmentation Strategy}

In M2CURL, data augmentation is crucial for multimodal contrastive learning. We specifically utilize random cropping and normalization, as these methods have shown to be highly effective \cite{DBLP:journals/corr/abs-2004-04136}. Random cropping (\( \mathcal{A}_{crop} \)) introduces variations in the input data by extracting different regions of the images, thereby improving the model's robustness to changes in perspective and scale. Normalization (\( \mathcal{N} \)), on the other hand, standardizes the pixel values across the dataset, which has been found to enhance learning stability and performance. The augmentation process for both visual (\( o_V \)) and tactile (\( o_T \)) observations is formulated as:

\[ o'_V = \mathcal{N}(\mathcal{A}_{crop}(o_V)), \quad o'_T = \mathcal{N}(\mathcal{A}_{crop}(o_T)) \]

where \( o'_V \) and \( o'_T \) are the augmented visual and tactile observations, respectively. This specific combination of augmentation techniques ensures a diverse range of perspectives and scales in the training data, which is key for learning invariant and robust features from visual data.

\subsection{Multimodal Contrastive Loss}

The M2CURL framework employs a sophisticated contrastive loss function that integrates both intra-modal and inter-modal learning, leveraging the specialized encoder heads for each modality.

\textbf{Intra-modal Contrastive Learning:}
For intra-modal learning, contrastive loss is computed separately for each modality:
\begin{itemize}
        \item \textit{Vision-to-Vision Contrastive Loss (\(\mathcal{L}_{VV}\))}: This loss is calculated using the vision-to-vision head \( H_{vv} \). It contrasts different augmented versions of the same visual observation to enhance feature learning within the visual domain.
        \item \textit{Tactile-to-Tactile Contrastive Loss (\(\mathcal{L}_{TT}\))}: Similarly, this loss uses the tactile-to-tactile head \( H_{tt} \) to contrast different tactile observations, fostering feature extraction in the tactile domain.
\end{itemize}

The loss for each modality is computed using the InfoNCE formula:
\begin{equation}
 \mathcal{L}_{VV/TT} = -\log \frac{\exp(\text{sim}(c_{vv/tt}(z_i), c_{vv/tt}(z_j)) / \tau)}{\sum_{k=1}^{N} \exp(\text{sim}(c_{vv/tt}(z_i), c_{vv/tt}(z_k)) / \tau)} \nonumber
\end{equation}
 
\textbf{Inter-modal Contrastive Learning:}
Inter-modal learning involves aligning the feature spaces across the visual and tactile modalities:
\begin{itemize}
    \item \textit{Vision-to-Tactile Contrastive Loss (\(\mathcal{L}_{VT}\))}: This part of the loss function uses the vision-to-tactile head \( H_{vt} \) to learn cross-modal representations, focusing on understanding the correlation between visual features and tactile sensory data.
    \item \textit{Tactile-to-Vision Contrastive Loss (\(\mathcal{L}_{TV}\))}: Similarly, this loss employs the tactile-to-vision head \( H_{tv} \) for learning cross-modal representations, aimed at comprehending how tactile features correlate with visual information.
\end{itemize}

The inter-modal contrastive loss is also based on the InfoNCE loss formula, adapted for cross-modal comparisons:
\begin{equation}
\mathcal{L}_{VT/TV} = -\log \frac{\exp(\text{sim}(c_{vt/tv}(z_i), c_{tv/vt}(z_j)) / \tau)}{\sum_{k=1}^{N} \exp(\text{sim}(c_{vt/tv}(z_i), c_{tv/vt}(z_k)) / \tau)} \nonumber
\end{equation}

\textbf{Combined Contrastive Loss:}
The combined multimodal contrastive loss in M2CURL, \( \mathcal{L}_{MM} \), integrates the intra-modal and inter-modal losses, each modulated by a specific balancing coefficient. This formulation allows for tailored learning from both within-modality and across-modality representations.

The combined loss is defined as:
\begin{equation}
 \mathcal{L}_{MM} = \lambda_{VV} \mathcal{L}_{VV} + \lambda_{TT} \mathcal{L}_{TT} + \lambda_{VT} \mathcal{L}_{VT} + \lambda_{TV} \mathcal{L}_{TV} \nonumber
\end{equation}

where $\lambda_{VV}$, $\lambda_{TT}$, $\lambda_{VT}$, and $\lambda_{TV}$ are the coefficients for vision-to-vision, tactile-to-tactile, vision-to-tactile, and tactile-to-vision learning, respectively.
% \begin{itemize}
%     \item \( \lambda_{VV}\) for vision-to-vision learning, controlling intra-modal learning in the visual domain.
%     \item \( \lambda_{TT} \) for tactile-to-tactile learning, adjusting intra-modal learning in the tactile domain.
%     \item \( \lambda_{VT} \) for vision-to-tactile learning, correlating visual features with tactile data.
%     \item \( \lambda_{TV} \) for tactile-to-vision learning, integrating tactile features into visual understanding.
% \end{itemize}
%
These coefficients enable precise control over the learning process, ensuring a balanced approach to multimodal representation integration and enhancing RL agent performance in diverse sensory environments.

\subsection{Integration with Reinforcement Learning}

The M2CURL framework's integration with RL is versatile, accommodating various RL architectures, including those beyond the traditional actor-critic paradigm. The key component, the multimodal contrastive loss (\( \mathcal{L}_{MM} \)), is adaptable to the specific RL setup in use.

For actor-critic architectures, \( \mathcal{L}_{MM} \) is integrated into the actor's loss function to enhance policy learning:
\begin{equation}
\mathcal{L'}_{actor} = \mathcal{L}_{actor} + \beta \mathcal{L}_{MM} \nonumber
\end{equation}
In this setup, \( \mathcal{L}_{actor} \) denotes the standard RL loss of the actor, and \( \beta \) acts as a weighting factor for the multimodal contrastive loss. Additionally, in such architectures, the critic's encoder is periodically updated by copying the weights from the actor's encoder. This ensures consistency between the policy evaluation by the critic and the policy updates by the actor.

In cases where an actor-critic algorithm is not employed, the integration of \( \mathcal{L}_{MM} \) adapts similarly to the specific algorithmic structure. For instance, the primary loss function of the RL algorithm can be modified to include \( \mathcal{L}_{MM} \), thereby infusing the learning process with multimodal data insights. This flexibility allows M2CURL to enhance a wide range of RL approaches, improving learning efficiency and policy performance across diverse environments and tasks.

\section{Experimental Framework}

\subsection{Evaluation Metrics}
In evaluating our methods and baselines, we focus on data efficiency and overall performance at two key milestones: 100k and 500k environment steps. This approach allows us to assess both the speed of initial learning and the asymptotic performance. We use two primary metrics for evaluation:

    \begin{enumerate}
        \item Sample Efficiency: Measured by the number of steps required by the baselines to match the performance of M2CURL at fixed environment steps (100k or 500k).
        \item Performance: Assessed by comparing the mean cumulative reward per episode.
    \end{enumerate}

\subsection{Environments}
The experiments are conducted in three Tactile Gym 2.0 environments, each chosen for its distinct challenges and complexity, providing a comprehensive testing ground for our algorithms.

\begin{enumerate}
    \item \textbf{\textit{Object Push}}: This environment involves the task of pushing a cube object along a trajectory generated by OpenSimplex Noise, emphasizing the need for precise manipulation and adaptability to unpredictable paths.
    \item \textbf{\textit{Edge Follow} }: The task here is to traverse a flat edge randomly oriented within a 360-degree range. The challenge is in keeping the edge centered on the sensor, requiring high precision and adaptability.
    \item \textbf{\textit{Surface Follow V2})}: A verticalized version of the \textit{Surface Follow V2} environment designed for training 4-Degree-of-Freedom (4-DoF) robots. This variant adds complexity by altering the orientation and dynamics of navigation, testing the robot's ability to adapt to vertical surfaces and different gravitational dynamics.
\end{enumerate}

These environments are ideal for evaluating the performance of M2CURL)in robotic manipulation tasks. They offer a thorough assessment of the algorithms in terms of precision, complexity, and coordination of different modalities to complete the task.

\subsection{Baselines for Benchmarking Sample Efficiency}
In our benchmarking process, M2CURL's performance was evaluated alongside SAC and PPO, including their versions augmented with RAD(Random Augmentation for Data-efficiency) ~\cite{laskin_lee2020rad}, a technique originally developed for visual tasks and here adapted for visuotactile perception. This comparative analysis aims to showcase the advancements M2CURL introduces in terms of learning convergence within tactile-rich environments. While comparing against the original versions of both RL algorithms, we want to rule out the efficacy that data augmentations may have on learning efficient representations of the visuotactile observations. Therefore, we implement the same augmentation approach used in M2CURL to preprocess the observations before feeding them into the RAD versions of the two RL algorithms. This comparison aims to explore whether RAD's augmentation techniques bolster SAC and PPO in multimodal environments and how M2CURL fares in harnessing visuotactile information for learning.

% Through this comprehensive benchmarking against both standard and augmented versions of well-established reinforcement learning algorithms, we aim to demonstrate M2CURL's distinctive strengths in learning efficiency and adaptability in environments rich with tactile feedback. The outcomes of these comparisons are anticipated to provide insightful conclusions regarding M2CURL's applicability in practical robotic manipulation tasks.

\subsection{Implementation Details}

M2CURL was implemented using Stable Baselines 3 \cite{stable-baselines3} and was tested with a simulated UR5 robot equipped with a DIGIT tactile sensor, using TCP velocity control. For the \textit{Object Push} and \textit{Surface Follow V2} environments, we used random trajectories, and for the \textit{Edge Follow}  variable vertical distances. A dense reward structure was applied for effective learning feedback. The modality weights ($\lambda_{VV, TT, VT, TV}$) were uniformly set to 1, with the contrastive loss weight ($\beta$) adjusted differently for SAC ($0.1$) and PPO ($1$) to address overfitting issues observed in SAC due to sample diversity. A higher temperature parameter ($\tau$) was used for SAC(0.1) than in PPO ($0.05$) to prevent overfitting of \( \lambda_{MM}\). This choice was made to address SAC's limited sample diversity. This adjustment softens the distribution over sample pairs, ensuring that even less similar samples get some positive probability. Consequently, it reduces the stark contrast in scores between positive and negative pairs, helping to mitigate overfitting.
Other hyperparameters and network architectures followed the Tactile Gym 2 simulator \cite{tactilegym2} settings, with heads comprising two fully connected layers with a hidden dimension of 2048. 

% \caption{Comparative Performance Analysis of M2CURL with Baseline RL Algorithms at 100K and 500K Training Steps, Including Standard Deviations. This table showcases the effectiveness of M2CURL, with performance scores (mean±std averaged over 3 runs) in the \textit{Object Push}, \textit{Edge Follow} , and \textit{Surface Follow V2}  environments, demonstrating improved sample efficiency and overall performance. Results for both SAC and PPO frameworks, including their standard, RAD-augmented, and state versions, are presented. Best performance in \textbf{bold.}}

\begin{table*}[h]
\centering
\caption{Comparative Performance Analysis of M2CURL with Baseline RL Algorithms at 100K and 500K Training Steps. This table showcases the effectiveness of M2CURL, with performance scores (mean±std averaged over 3 runs) in the \textit{Object Push}, \textit{Edge Follow}, and \textit{Surface Follow V2}  environments, demonstrating improved sample efficiency and overall performance. Results for both SAC and PPO frameworks, including their standard, RAD-augmented, and state versions, are presented. Best performance in \textbf{bold.}}
\label{tab:your_table_label}
\begin{tabular}{lcccc|cccc}
\hline
\textbf{500K Steps} & \textbf{\textcolor{blue}{M2CURL SAC}} & RAD SAC & \multicolumn{1}{c:}{SAC} & \multicolumn{1}{c||}{SAC-state} &\textbf{ \textcolor{blue}{M2CURL PPO}} & RAD PPO & \multicolumn{1}{c:}{PPO} & \multicolumn{1}{c}{PPO-state} \\ \hline
\textit{Object Push}     & \textbf{-63.3±4.3}    & -94.0±12.5           & \multicolumn{1}{c:}{-64.8±8.2}       & \multicolumn{1}{c||}{-83.1±12.6}              & \textbf{-67.5±2.0}     & -162.0±3.3          & -187.2±3.1      & -18.8±1.7              \\
\textit{Edge Follow}      & \textbf{-24.1±1.3}    & -27.3±1.4           & \multicolumn{1}{c:}{-27.2±4.5}       & \multicolumn{1}{c||}{-35.5±3.8}              & \textbf{-15.6±2.1 }    & -19.0±2.3           & \multicolumn{1}{c:}{-21.6±4.6}       & -30.5±5.0              \\ 
\textit{Surface Follow V2} & \textbf{-37.4±2.1}  & -46.8±5.1           & \multicolumn{1}{c:}{-178.6±24.1}      & \multicolumn{1}{c||}{-39.7±5.01}              & \textbf{-48.0±2.3}     & -64.2±2.8           & \multicolumn{1}{c:}{-61.4±2.6}       & -8.9±1.5               \\ \hline
\textbf{100K Step Scores} & & & \multicolumn{1}{c:}{} & \multicolumn{1}{c||}{} \\ \hline
\textit{Object Push}     & \textbf{-114.3±13.4}   & -181.9±21.9          & \multicolumn{1}{c:}{-190.6±14.1}      & \multicolumn{1}{c||}{-111.6±13.5}             & \textbf{-179.7±11.6}    & -183.4±19.8          & \multicolumn{1}{c:}{-237.6±14.5}      & -52.3±4.2              \\
\textit{Edge Follow}      & \textbf{-31.0±0.9}    & -38.9±1.9           & \multicolumn{1}{c:}{-42.8±2.8}       & \multicolumn{1}{c||}{-35.5±3.7}              & \textbf{-14.7±1.2}     & -17.7±2.4           & \multicolumn{1}{c:}{-32.6±1.8}       & -44.8±4.1              \\ 
\textit{Surface Follow V2} & \textbf{-45.1±3.2}  & -78.6±6.1           & \multicolumn{1}{c:}{-63.8±5.5}       & \multicolumn{1}{c||}{-60.3±4.6}              & \textbf{-6.5±0.7 }     & -23.7±1.1           & \multicolumn{1}{c:}{-18.9±2.6}       & -16.6±1.4              \\ \hline
\end{tabular}
\end{table*}

\section{Results}
Our experimental evaluation assesses the performance of the M2CURL framework against the SAC and PPO frameworks, in their standard and RAD-augmented versions. 
We also compare against the same algorithms with access to the actual state of the robot instead of the visuotactile observations. 
The different milestones at 100K and 500K timesteps were chosen to understand the initial adaptation and the more mature phase of learning of each algorithm, providing insights into both the early learning phase (100k steps) and the more mature phase of learning (500k steps).

In the \textit{\textit{Object Push}} task, M2CURL demonstrated a significant advantage over the baselines. At 500k steps, M2CURL achieved higher scores for both SAC and PPO frameworks. Notably, M2CURL showed its rapid adaptation capability in the early learning phase, outperforming others at 100k steps, indicative of its enhanced sample efficiency. The effectiveness of the multimodal contrastive loss becomes more apparent by comparing the performance of the M2CURL PPO algorithm with the RAD SAC and SAC algorithms. In general, off-policy algorithms, like SAC, are more sample-efficient than on-policy algorithms, like PPO. This is apparent by comparing the performance of the two algorithms in Table \ref{tab:your_table_label}. Despite that, M2CURL PPO manages to converge faster than the RAD SAC and SAC, highlighting the effectiveness of our method.
M2CURL's sample efficiency is consistent in the 'Edge Following' task too, outpacing both the standard and RAD-augmented versions of either RL framework.

\begin{table}[h]
\centering
\caption{Ablation study in the 'Object-Push' environment, comparing the performance of SAC when trained exclusively with intra-modality losses, exclusively with inter-modality losses, and the original M2CURL loss.}
\label{tab:ablation}
\begin{tabular}{lccc}
\hline
\textbf{Steps} & \textbf{M2CURL} & \textbf{Intra-Modality} & \textbf{Inter-Modality} \\ \hline
\textbf{500K} & -63.37±6.4 & -140.45±19.3 & -63.73±8.7 \\ 
\hline
\textbf{100K} & -114.32±16.6 & -154.76±13.8 & -261.40±45.2 \\
\hline
\end{tabular}
\end{table}

In the \textit{Surface Follow V2}  task, M2CURL notably mitigated the divergence that occurred in the later stages of training the PPO instances. This task's complexity, requiring navigation and adaptation to vertical surfaces and varied dynamics, typically poses challenges for on-policy algorithms like PPO. However, the multimodal contrastive learning approach in M2CURL played a crucial role in ensuring learning stability and efficiency, demonstrating its robustness and adaptability in this challenging environment.

The results across both tasks and at both the early (100k) and later (500k) stages of learning highlight M2CURL's sample efficiency and long-term performance capabilities. 
The outcomes observed at the 100K mark are especially indicative, as they demonstrate M2CURL's ability to quickly grasp and adapt to new tasks, a crucial aspect in dynamic environments where rapid learning is essential. The consistent high-level performance of M2CURL in later stages, coupled with its ability to outperform even algorithms with direct access to the state, highlights the framework's proficiency in learning efficient visuotactile representations.

\begin{figure}[htbp]
    \includegraphics[width=\linewidth]{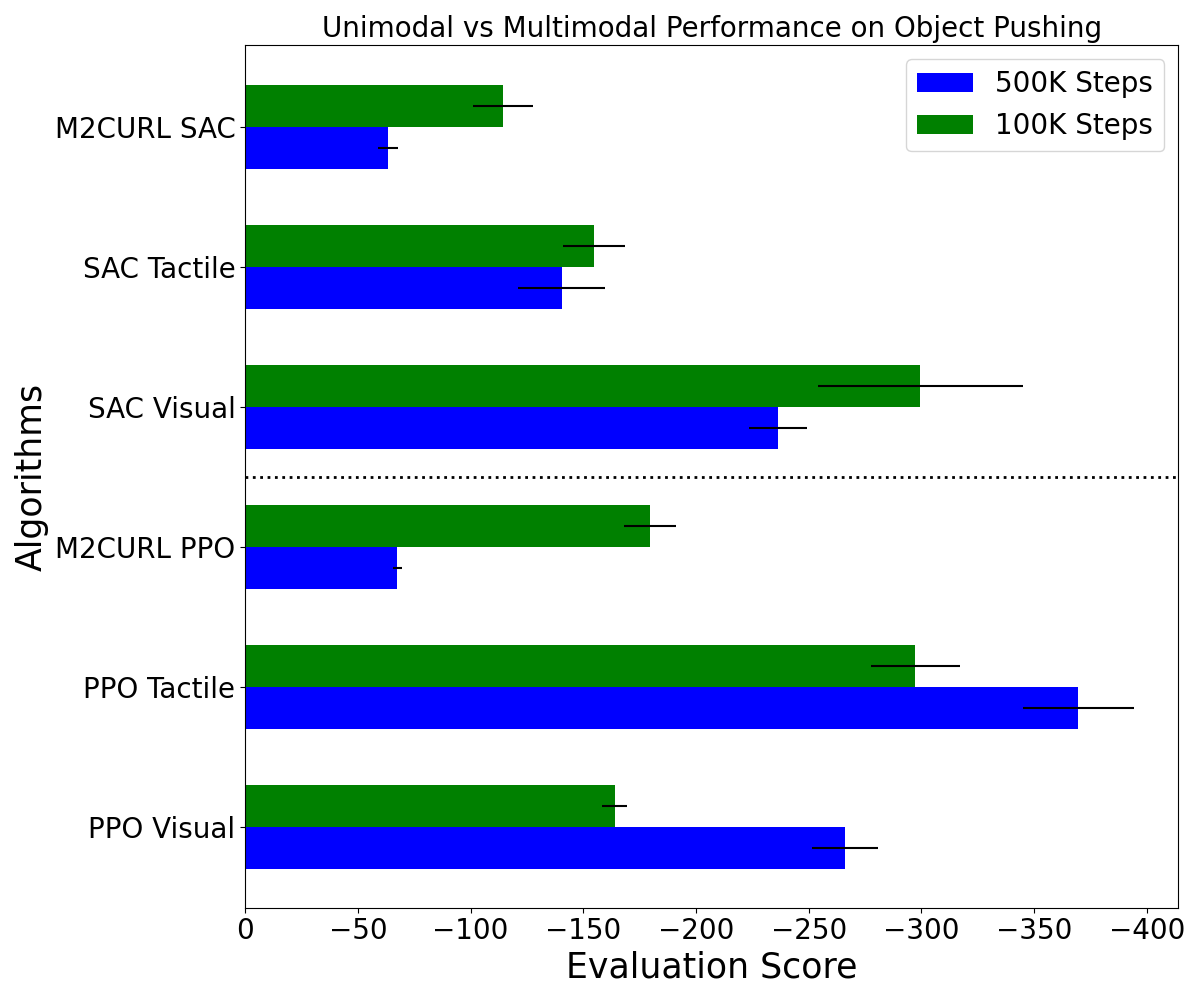}
    \caption{Performance comparison of SAC and PPO algorithms using visual or tactile observations, against M2CURL using visuotactile observations.}
    \label{fig:plot}
\end{figure}
To further assess M2CURL's performance relative to a simpler contrastive loss framework, we conducted ablation studies detailed in Table \ref{tab:ablation}. In these studies, the SAC algorithm was trained in the 'Object-Push' environment using only intra-modality losses (\( \lambda_{VT} = \lambda_{TV} = 0 \)) and separately with only inter-modality losses (\( \lambda_{VV} = \lambda_{TT} = 0 \)). The results, as presented in the table, unequivocally demonstrate that a combined approach of both intra and inter-modality losses within M2CURL leads to superior visuotactile representation learning, underscoring the framework's advanced capability.

Finally, we aimed to highlight the importance of modality fusion in tactile-rich settings. In our study shown in Figure \ref{fig:plot}, we compared unimodal (visual-only and tactile-only) reinforcement learning against the multimodal approach in M2CURL. The results clearly showed that M2CURL's integration of both visual and tactile data significantly outperforms unimodal methods for either RL algorithm. This underscores the effectiveness of multimodal learning in complex environments, demonstrating the advantage of combining modalities for improved reinforcement learning performance.

% \begin{table}[h]
% \centering
% \caption{Performance Scores for SAC and PPO with Visual and Tactile Observations}
% \label{tab:performance_scores}
% \begin{tabular}{lcccc}
% \hline
%  & \multicolumn{2}{c}{\textbf{SAC}} & \multicolumn{2}{c}{\textbf{PPO}} \\
% \textbf{Steps} & \textbf{Visual} & \textbf{Tactile} & \textbf{Visual} & \textbf{Tactile} \\
% \hline
% \textbf{500K} & -63.37±6.4 & -140.45±19.3 & -266.07±14.63 & -369.57±24.64 \\
% \hline
% \textbf{100K} & -114.32±16.6 & -154.76±13.87 & -163.90±5.55 & -297.24±19.70 \\
% \hline
% \end{tabular}
% \end{table}

% \section{Ablation study}
% \section{Discussion}

\section{Conclusions}

In this paper, we introduced M2CURL, a novel framework that leverages a multimodal contrastive loss to enhance the efficiency of RL agents in tactile-rich robotic manipulation tasks. M2CURL processes both inter-modality and intra-modality losses, facilitating the learning of efficient representations. These representations are then utilized by the actor and critic components of two state-of-the-art RL algorithms, an on-policy and an off-policy one. Our empirical results underscore the substantial benefits M2CURL offers compared to the simple representation concatenation for the two modalities in RL algorithms.
% Notably, M2CURL excels in both early learning efficiency and long-term performance, which is crucial for practical applications in dynamic and unpredictable environments.
%
% We proved that the integration of the multimodal contrastive loss within M2CURL, leveraging self-supervised learning techniques, has proven to be particularly effective. 
% The multimodal contractive loss approach enables the framework to efficiently integrate and learn from high-dimensional, multimodal sensory data, which is a significant advancement in the field of reinforcement learning. 
M2CURL's algorithm-agnostic nature further enhances its utility, allowing for easy integration with various existing RL algorithms.
Our findings suggest that M2CURL can play a pivotal role in advancing robotic manipulation tasks, especially those requiring the integration of complex and varied sensory inputs. The framework's ability to rapidly adapt and maintain robust performance over time offers new opportunities for deploying more intelligent and capable robotic systems in a wide range of real-world applications.
Future work will focus on extending the application of M2CURL to physical robotic systems and exploring its scalability and effectiveness in more diverse environments and with additional sensory modalities. 
% This will further our understanding of the framework's potential and pave the way for its broader adoption in advanced robotic applications.



% \section*{ACKNOWLEDGMENT}

% The preferred spelling of the word ÒacknowledgmentÓ in America is without an ÒeÓ after the ÒgÓ. Avoid the stilted expression, ÒOne of us (R. B. G.) thanks . . .Ó  Instead, try ÒR. B. G. thanksÓ. Put sponsor acknowledgments in the unnumbered footnote on the first page.

% %%%%%%%%%%%%%%%%%%%%%%%%%%%%%%%%%%%%%%%%%%%%%%%%%%%%%%%%%%%%%%%%%%%%%%%%%%%%%%%%

% References are important to the reader; therefore, each citation must be complete and correct. If at all possible, references should be commonly available publications.

\printbibliography

\end{document}